\documentclass{article}
\usepackage[breaklinks=true,bookmarks=false]{hyperref}
\usepackage{spconf,amsmath,graphicx}
\usepackage{float}
\usepackage{color}
\usepackage{subcaption}
\usepackage{graphicx}

\title{ShuffleSeg: Real-time Semantic Segmentation  Network}

 \name {Mostafa Gamal, Mennatullah Siam, Mo'men Abdel-Razek}
\address{mostafa.gamal95@eng-st.cu.edu.eg, mennatul@ualberta.ca, \\ moemen.abdelrazek96@eng-st.cu.edu.eg \\ Cairo University, University of Alberta}

\begin{document}
%\ninept
%
\maketitle

\begin{abstract}
Real-time semantic segmentation is of significant importance for mobile and robotics related applications. We propose a computationally efficient segmentation network which we term as ShuffleSeg. The proposed architecture is based on grouped convolution and channel shuffling in its encoder for improving the performance. An ablation study of different decoding methods is compared including Skip architecture, UNet, and Dilation Frontend. Interesting insights on the speed and accuracy tradeoff is discussed. It is shown that skip architecture in the decoding method provides the best compromise for the goal of real-time performance, while it provides adequate accuracy by utilizing higher resolution feature maps for a more accurate segmentation.  ShuffleSeg is evaluated on CityScapes and compared against the state of the art real-time segmentation networks. It achieves 2x GFLOPs reduction, while it provides on par mean intersection over union of 58.3\% on CityScapes test set. ShuffleSeg runs at 15.7 frames per second on NVIDIA Jetson TX2, which makes it of great potential for real-time applications.
\end{abstract}
\section{Introduction}
Building computationally efficient convolutional neural networks (CNNs) is still an open research problem. There has been two main mechanisms for improving their efficiency. The first mechanism is focused on designing efficient models, such as the work in GoogleNet \cite{szegedy2015going}, Xception \cite{chollet2016xception}, MobileNet \cite{howard2017mobilenets}, and the recent ShuffleNet \cite{zhang2017shufflenet}. The other mechanism is directed towards model acceleration, by pruning network connections \cite{han2015deep}\cite{han2015learning} or channels \cite{wen2016learning} or network quantization\cite{rastegari2016xnor}\cite{wu2016quantized}. Previous work in improving computational efficiency mostly focused on the end task of image classification and object detection. However, few works were targeted towards real-time semantic segmentation networks although semantic segmentation has numerous benefits in robotics related applications \cite{cordts2016cityscapes}\cite{xu2017end}\cite{wong2017segicp}. That highlights the need for computationally efficient semantic segmentation.

In this paper, we propose a real-time semantic segmentation network based on the ShuffleNet unit introduced in \cite{zhang2017shufflenet}. We refer to this network as ShuffleSeg throughout the paper. ShuffleSeg incorporates skip connections in its decoder for improved segmentation results. Our proposed network requires 2.03 GFLOPs which outperforms the state of the art in computationally efficient segmentation networks that required 3.83 GFLOPs\cite{paszke2016enet}. Nonetheless, ShuffleSeg achieves comparable mean intersection over union of 58.2\% on CityScapes test set benchmark. Thus, our network provides good balance between speed and accuracy. This provides potential benefits for further deployment on embedded devices.

Real-time semantic segmentation started to draw attention recently. Paszke et. al. \cite{paszke2016enet} introduced ENet as an efficient lightweight segmentation network with a bottleneck module. Chaurasia et. al. \cite{chaurasia2017linknet} proposed LinkNet architecture that uses ResNet18 as its encoder. LinkNet achieves better mean intersection over union than ENet. However, ENet outperforms it in terms of computational efficiency. Other networks not focused on computational efficiency, but are widely used in segmentation literature, are SegNet and FCN8s. Badrinarayanan et. al. \cite{badrinarayanan2015segnet} proposed SegNet as an early attempt for end-to-end semantic segmentation in encoder-decoder architecture. Long et. al. \cite{long2015fully} proposed the first attempt for fully convolutional segmentation network (FCN) that was trained end to end. He also proposed the skip-net method to utilize higher resolution feature maps in the segmentation in FCN16s and FCN8s architectures.

To the best of our knowledge, no previous work on real-time semantic segmentation utilized group convolution and channel shuffling. In this work, we propose ShuffleSeg as a computationally efficient segmentation network. The network is inspired from ShuffleNet \cite{zhang2017shufflenet} which provided an efficient classification and detection networks. The ShuffleNet unit uses grouped convolution instead of 1x1 convolution to boost the performance. Grouped convolution alone can hurt the network accuracy, therefore channel shuffling is used to maintain good accuracies. This is coupled with skip architecture to improve our segmentation results, by using higher resolution feature maps. The code for ShuffleSeg will be publicly available at \footnote{https://github.com/MSiam/TFSegmentation}. 
%The paper is organized as follows: section \ref{sec:method} describes the detailed network architecture for ShuffleSeg, section \ref{sec:exps} provides experimental analysis and ablation study of different design choices. Finally, section\ref{sec:conc} presents concluding remarks.  

\section{Method}
In this section, a detailed description of the network architecture used for semantic segmentation is presented. The architecture is explained as two main modules; the encoding module is the one responsible for extracting features, while the decoding module is responsible for with-in the network up-sampling to compute the final class probability maps.

\subsection{Encoder Architecture} 
The encoder used in our proposed architecture is based on ShuffleNet \cite{zhang2017shufflenet}. We primarily inspire from their grouped convolution and channel shuffling. It was shown in \cite{zhang2017shufflenet}\cite{chollet2016xception}\cite{howard2017mobilenets} that depthwise separable convolution or grouped convolution reduce the computational cost, while maintaining good representation capability. Stacking grouped convolution can lead to one major bottleneck. The output channels are going to be derived from a limited fraction of input channels. To solve such an issue channel shuffling was introduced in \cite{zhang2017shufflenet} and is used as well inside ShuffleSeg architecture. 

\begin{figure}[ht!]
\centering
\includegraphics[scale= 0.24]{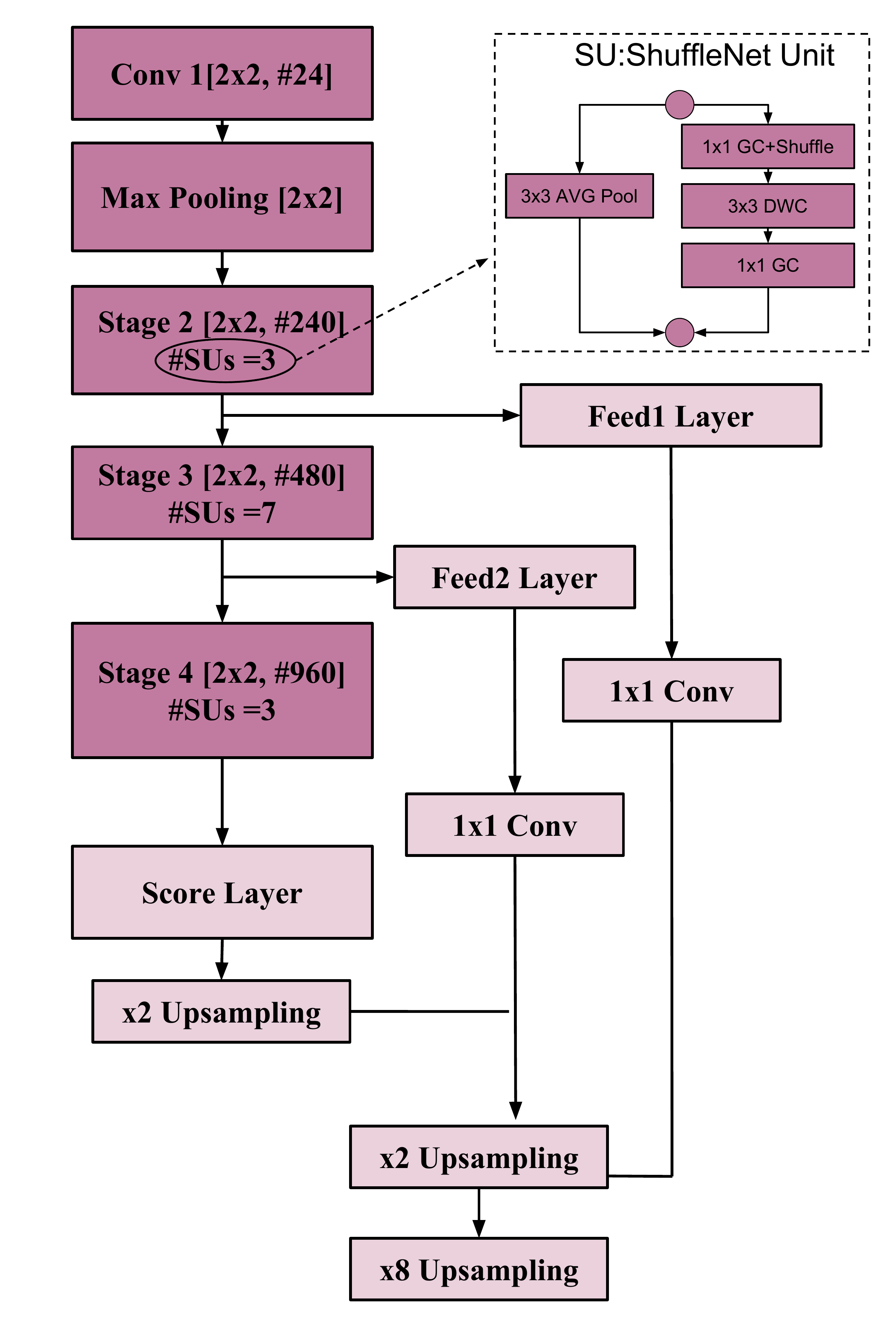}
\caption{ShuffleSeg Network Architecture.}
\label{fig:shuffleseg}
\end{figure}

%We have selected three groups to be used in the grouped convolution. This is largely based on the experiments presented by \cite{zhang2017shufflenet}, that showed that three groups provided a good compromise between accuracy and efficiency.
Figure \ref{fig:shuffleseg} shows the network architecture of ShuffleSeg for both encoding and decoding parts. An initial 3x3 convolutional layer is used with stride 2 for downsampling, followed by 2x2 maxpooling. Then, three stages are used, each stage is composed of multiple ShuffleNet units. Stage 2 and 4 are composed of 3 ShuffleNet units while stage 3 has 7 units. This leads to a down-sampling factor of resolution 32.

ShuffleNet units act as residual bottleneck modules. On the primary branch, it has average pooling with stride 2. However, on the residual branch it has 1x1 grouped convolution then depth-wise convolution with stride 2 then another 1x1 grouped convolution. The purpose of the first grouped convolution is to act as a bottleneck to reduce the number of channels for a computationally efficient solution. This is followed by shuffling channels to ensure that all the input are connected to the output without restrictions from assigned groups. The second grouped convolution recovers the channels back to the number of input channels for a better representation capability. Figure \ref{fig:shuffleseg} shows the detailed structure of this ShuffleNet unit in the upper right. 

\subsection{Decoder Architecture} 
Transposed convolutions are performed in the decoding part of the segmentation network in order to upsample to the input resolution. Different decoding methods are used which inspire from the work in UNet \cite{DBLP:journals/corr/RonnebergerFB15}, FCN8s \cite{long2015fully} and Dilation Frontend\cite{yu2015multi}. Four different decoding methods are compared which are: (1) UNet. (2) SkipNet. (3) Dilation Frontend 8s. (4) Dilation 4s. Insights regarding how these methods affect the network efficiency and accuracy are provided in the ablation study section \ref{sec:ablation}. To the best of our knowledge, this is the first study to provide such insights. This comparison would benefit the research community and boost the direction toward real-time segmentation further on. 

\textbf{SkipNet:} The main idea in skip connections\cite{long2015fully} is to benefit from higher resolution feature maps to improve accuracy. It is worth noting that transposed convolutions in this decoding method are applied on the final heat maps with channels equivalent to the required classes. This ensures the computational efficiency of the network, as the number of channels have direct impact on it. The upsampling factor required is of resolution 32. In our case, the output from stage 4 is passed through 1x1 convolution, denoted as score layer, to convert the channels to the number of classes. Stage 2 and stage 3 are used as input intermediate layers to improve the heat map resolution. The output from score layer is upsampled by 2x and then elementwise addition between these and heatmaps from stage 3 is performed. An equivelant operation is performed for stage 2. Finally, a transposed convolution of stride 8 provides the final probability maps that match the input size. Transposed convolutions are initialized with bilinear upsampling.

\textbf{UNet:} UNet decoding method creates an up-sampling stage corresponding to each downsampling in the original network. Feature fusion is then performed using elementwise addition with the corresponding stage. Hence, feature concatenation could be used for the fusion method, but elementwise addition provides a computationally efficient solution that better serves the realtime perspective. It is worth noting that transposed convolutions in this method are performed on feature maps instead of heat maps as in SkipNet. This leads to an increase in the number of channels that hurts the computational efficiency in comparison to SkipNet.

\textbf{Dilation Frontend 8s and 4s:} In this decoding method similar to \cite{yu2015multi}, the ShuffleNet downsampling factor is changed from 32 to 8. This is done by changing the final units with stride 2 to 1. Dilation convolution with dilation rate of 2 then 4 is used, to maintain the higher receptive field. Then, a transposed convolution of stride 8 is performed to provide the final probability map, this denotes Dilation 8s. As for Dilation 4s, the same architecture is used while replacing the last transposed convolution with stride 4 and downsampling the target labels during training by factor of 2. Downsampling the labels with a small factor turned to provide a more efficient method while keeping adequate accuracy in comparison to the original resolution.

\label{sec:method}

\section{Experiments}
\begin{figure*}[ht!]
\centering
\begin{subfigure}{0.43\textwidth}
    \includegraphics[scale= 0.21]{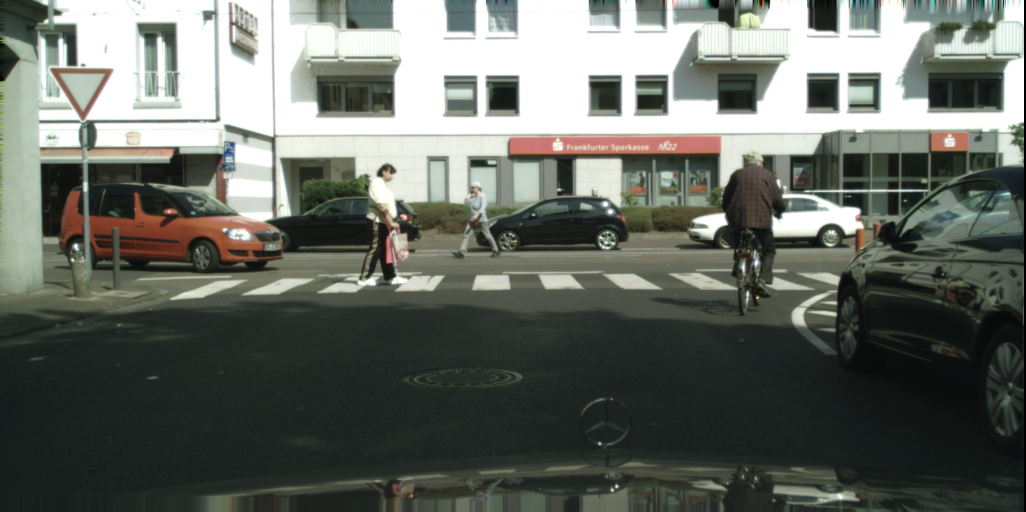}
    \caption{}
\end{subfigure}%
\begin{subfigure}{0.43\textwidth}
    \includegraphics[scale= 0.21]{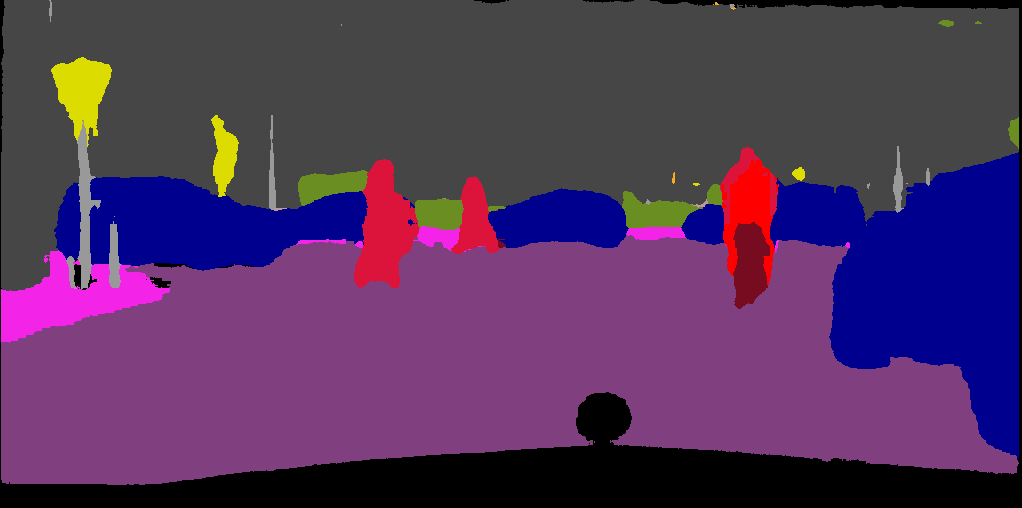}
    \caption{}
\end{subfigure}

\begin{subfigure}{0.43\textwidth}
    \includegraphics[scale= 0.21]{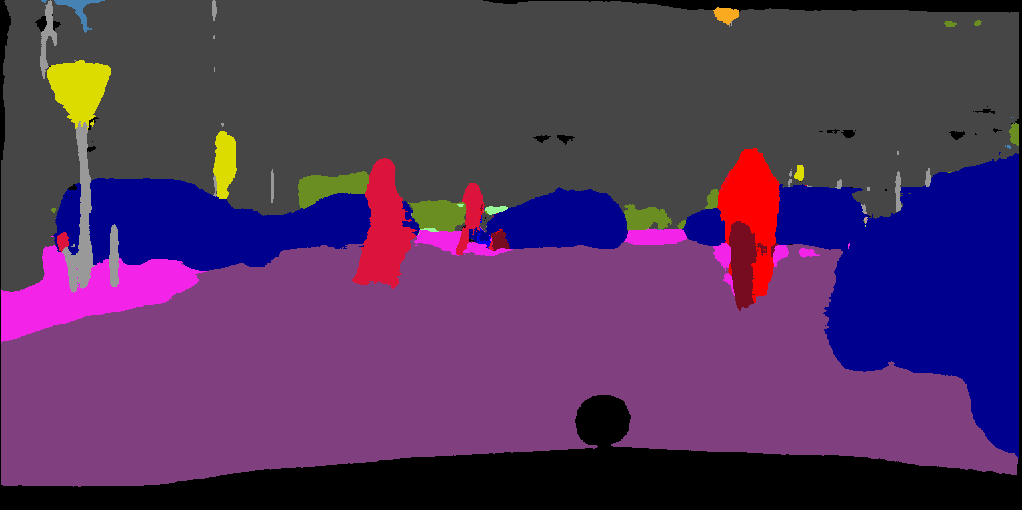}
    \caption{}
\end{subfigure}%
\begin{subfigure}{0.43\textwidth}
    \includegraphics[scale= 0.21]{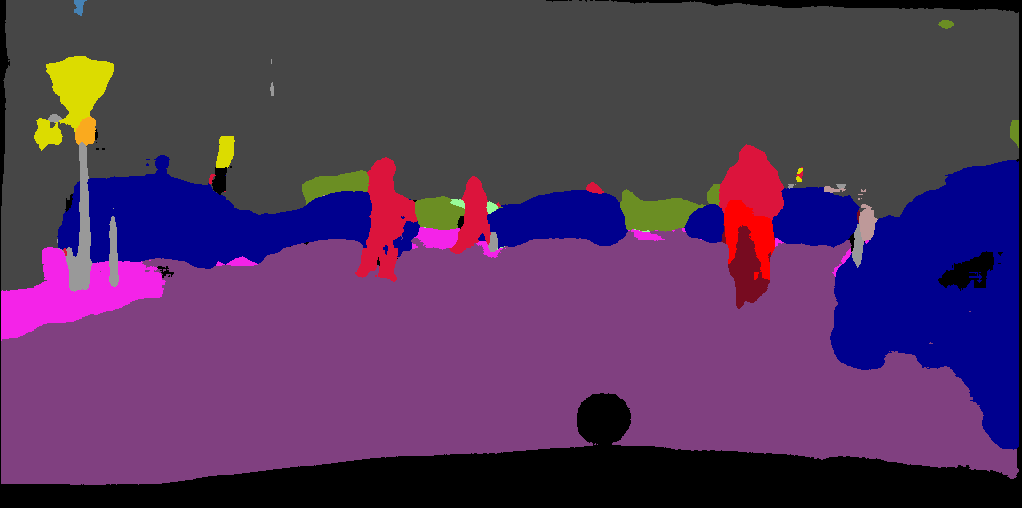}
    \caption{}
\end{subfigure}
\caption{ShuffleSeg Qualitative Images on CityScapes. (a) Original Image. (b) SkipNet pretrained with Coarse Annotations. (b) UNet. (c) Dilation 8s.}
\label{fig:qual_city}
\end{figure*}

\begin{table*}[ht!]
\centering
\caption{Comparison of different decoding methods in accuracy and computational efficiency on class level.}
\label{table:ablationclass}
\begin{tabular}{|l|l|l|l|l|l|l|l|l|l|l|l|l|}
\hline
Model & GFLOPs & mIoU & Building & Sky & Car & Sign & Road & Person & Fence & Pole & Sidewalk & Bicycle \\ \hline
UNet & 17.7 & 57.0 & 83.7 & 89.0 & \textbf{87.8} & 54.3 & 95.1 & 61.7 & 34.6 & 40.3 & 69.5 & 59.9 \\ \hline
SkipNet & \textbf{4.52} & 55.5 & 83.9 & 88.6 & 86.5 & 50.5 & 94.8 & 60.8 & 35.2 & 37.9 & 68.6 & 58.8 \\ \hline
Dilation8s & 17.7 & 53.9 & 84.1 & 90.3 & 86.6 & \textbf{57.3} & \textbf{95.2} & \textbf{62.9} & 31.4 & 37.2 & 68.5 & \textbf{60.2} \\ \hline
Dilation4s & 16 & 53.4 & 82.7 & 89.4 & 85.1 & 53.8 & 94.1 & 57.9 & 32.0 & 35.5 & 66.4 & 59.4 \\ \hline
Coarse & \textbf{4.52} & \textbf{59.3} & \textbf{85.5} & \textbf{90.8} & 87.5 & 54.9 & 94.6 & 60.2 & \textbf{41.7} & \textbf{40.8} & \textbf{70.5} & 58.8 \\ \hline
\end{tabular}
\end{table*}

\begin{table*}[ht!]
\centering
\caption{Comparison of different decoding methods in accuracy on category level.}
\label{table:ablationcategory}
\begin{tabular}{|l|l|l|l|l|l|l|l|l|}
\hline
Model & mIoU & Flat & Nature & Object & Sky & Construction & Human & Vehicle \\ \hline
UNet & 79.1 & 95.9 & 87.6 & 47.8 & 89.0 & 83.7 & 64.0 & 85.5 \\ \hline
SkipNet & 78.2 & 95.9 & 87.0 & 44.5 & 88.6 & 83.8 & 63.3 & 84.2 \\ \hline
Dilation8s & 79.3 & \textbf{96.6} & 87.3 & 45.4 & 90.3 & 84.4 & \textbf{66.4} & 84.7 \\ \hline
Dilation4s & 77.4 & 95.2 & 86.4 & 43.8 & 89.4 & 82.7 & 61.2 & 83.0 \\ \hline
Coarse & \textbf{79.4} & 94.8 & \textbf{87.8} & \textbf{48.1} & \textbf{90.8} & \textbf{85.3} & 63.4 & \textbf{85.5} \\ \hline
\end{tabular}
\end{table*}

\begin{table*}[ht!]
\centering
\caption{Comparison of ShuffleSeg to the state of the art real-time segmentation networks.}
\label{table:quant_city}
\begin{tabular}{|l|l|l|l|l|l|}
\hline
Model & GFLOPs & Class IoU & Class iIoU & Category IoU & Category iIoU \\ \hline
SegNet & 286.03 & 56.1 & \textbf{34.2} & 79.8 & \textbf{66.4} \\ \hline
ENet &  3.83 & 58.3 & 24.4 & \textbf{80.4} & 64.0 \\ \hline
%FCN8s & - & \textbf{65.3} & \textbf{41.7} & \textbf{85.7} & \textbf{70.1} \\ \hline
ShuffleSeg & \textbf{2.03} & \textbf{58.3} & 32.4 & 80.2 & 62.2 \\ \hline
\end{tabular}
\end{table*}

\subsection{Dataset}
Evaluation is conducted on CityScapes dataset\cite{cordts2016cityscapes}. It contains 5000 images with fine annotation, with 20 classes including the ignored class. 
The dataset is split into 2975 images for training, 500 for validation and 1525 for testing. It also contains another 20,000 images annotated but with coarse annotation. Coarse annotations are used for pretraining our final model that resulted in an improved accuracy as shown in the ablation study.

\subsection{Experimental Setup}
Experiments are conducted on images with size 512x1024, with 20 classes including the last class for the ignored class. A weighted cross entropy loss is used from \cite{paszke2016enet}, to overcome the imbalance in the data between different classes. The class weight is computed as $w_{class}=\frac{1}{ln(c+p_{class})}$, where $c$ is a constant hyper-parameter with value 1.02. L2 regularization is used to avoid over-fitting with weight decay rate of 5$e^{-4}$. Adam optimizer\cite{kingma2014adam} is used with learning rate 1$e^{-4}$. Batch normalization\cite{ioffe2015batch} is used after all convolutional or transposed convolution layers, to ensure faster convergence. The feature extractor part of the network is initialized with pre-trained ImageNet weights. Throughout all the experiments, we use 3 as the number of groups in grouped convolution. The code is built on TensorFlow and will be made public to the research community to benefit from it.

\subsection{Ablation Study}
\label{sec:ablation}
A detailed ablation study of different decoding methods is provided in this section. Four different decoding methods are compared, which are: (1) UNet. (2) SkipNet. (3) Dilation8s. (4) Dilation4s. Table \ref{table:ablationclass} and \ref{table:ablationcategory} shows the results for GFLOPs on images size 1024x512, mean IoU, perclass IoU, and percategory IoU on the validation set in cityscapes. It clearly demonstrates that SkipNet architecture provides the most efficient method while providing relatively good accuracy. Although UNet provides a more accurate solution, yet is not efficient for real-time solutions. Dilation 8s and 4s provide an inefficient solution that does not serve the real-time goal. They also suffer with classes that have fewer representation in the dataset, while outperform on more represented classes. This is shown in Table \ref{table:ablationcategory} as they outperform in mean IoU and in categories with higher representation such as Flat and Sky.

The above observations motivate the usage of SkipNet decoding method for further experiments on ShuffleSeg. In order to improve the network accuracy on classes with smaller representation than others in the dataset such as fence and pole, pretraining of the network is performed. Pretraining is conducted on the coarse annotation section with more labeled images. Afterwards, the network is trained on the fine annotations sections. This alone led to an improvement of 4\% in overall mean IoU. This is termed as Coarse in Table \ref{table:ablationclass} and provides mean IoU of 59.3\%.

\subsection{State of the Art Comparison}
Comparison to the state of the art real-time segmentation is provided in Table \ref{table:quant_city}. Experiments in this section are conducted on the test set. ShuffleSeg, which is the SkipNet version with coarse pretraining, outperforms ENet\cite{paszke2016enet} in computational efficiency. It lead to 2x reduction in GFLOPs, while it is on par with it in accuracy. In comparison to SegNet\cite{badrinarayanan2015segnet}, ShuffleSeg outperforms it in accuracy while having 141x GFLOPs reduction. This clearly shows the benefit of grouped convolution with channel shuffle on the reduction of operations required. ShuffleSeg runs at 15.7 frames per second on NVIDIA Jetson TX2, so it offers real-time performance on embedded devices.  Figure \ref{fig:qual_city} shows the qualitative results of ShuffleSeg on CityScapes. UNet provide a more detailed and accurate segmentation than Dilation 8s. The SkipNet architecture pretrained with coarse annotated data provide the most accurate segmentation in comparison to UNet and Dilation 8s. It is shown that ShuffleSeg can provide good accuracy with less number of operations. 
\label{sec:exps}

\section{Conclusion}
In this paper, an architecture based on grouped convolution and channel shuffling in its encoder is used. An ablation study is performed to compare different decoding methods. Interesting insights on the speed and accuracy trade-off is discussed. It is shown that skip architecture in the decoding method provides the best compromise between computational efficiency and accuracy. ShuffleSeg achieves 2x GFLOPs reduction in comparison to ENet and 141x in comparison to SegNet. It still provides on par mean intersection over union of 58.3\% on CityScapes test set. 
\label{sec:conc}

%%%%%%%%% BODY TEXT
\bibliographystyle{IEEEbib}
\bibliography{refs}

\begin{thebibliography}{10}

\bibitem{szegedy2015going}
Christian Szegedy, Wei Liu, Yangqing Jia, Pierre Sermanet, Scott Reed, Dragomir
  Anguelov, Dumitru Erhan, Vincent Vanhoucke, and Andrew Rabinovich,
\newblock ``Going deeper with convolutions,''
\newblock in {\em Proceedings of the IEEE conference on computer vision and
  pattern recognition}, 2015, pp. 1--9.

\bibitem{chollet2016xception}
Fran{\c{c}}ois Chollet,
\newblock ``Xception: Deep learning with depthwise separable convolutions,''
\newblock {\em arXiv preprint arXiv:1610.02357}, 2016.

\bibitem{howard2017mobilenets}
Andrew~G Howard, Menglong Zhu, Bo~Chen, Dmitry Kalenichenko, Weijun Wang,
  Tobias Weyand, Marco Andreetto, and Hartwig Adam,
\newblock ``Mobilenets: Efficient convolutional neural networks for mobile
  vision applications,''
\newblock {\em arXiv preprint arXiv:1704.04861}, 2017.

\bibitem{zhang2017shufflenet}
Xiangyu Zhang, Xinyu Zhou, Mengxiao Lin, and Jian Sun,
\newblock ``Shufflenet: An extremely efficient convolutional neural network for
  mobile devices,''
\newblock {\em arXiv preprint arXiv:1707.01083}, 2017.

\bibitem{han2015deep}
Song Han, Huizi Mao, and William~J Dally,
\newblock ``Deep compression: Compressing deep neural networks with pruning,
  trained quantization and huffman coding,''
\newblock {\em arXiv preprint arXiv:1510.00149}, 2015.

\bibitem{han2015learning}
Song Han, Jeff Pool, John Tran, and William Dally,
\newblock ``Learning both weights and connections for efficient neural
  network,''
\newblock in {\em Advances in Neural Information Processing Systems}, 2015, pp.
  1135--1143.

\bibitem{wen2016learning}
Wei Wen, Chunpeng Wu, Yandan Wang, Yiran Chen, and Hai Li,
\newblock ``Learning structured sparsity in deep neural networks,''
\newblock in {\em Advances in Neural Information Processing Systems}, 2016, pp.
  2074--2082.

\bibitem{rastegari2016xnor}
Mohammad Rastegari, Vicente Ordonez, Joseph Redmon, and Ali Farhadi,
\newblock ``Xnor-net: Imagenet classification using binary convolutional neural
  networks,''
\newblock in {\em European Conference on Computer Vision}. Springer, 2016, pp.
  525--542.

\bibitem{wu2016quantized}
Jiaxiang Wu, Cong Leng, Yuhang Wang, Qinghao Hu, and Jian Cheng,
\newblock ``Quantized convolutional neural networks for mobile devices,''
\newblock in {\em Proceedings of the IEEE Conference on Computer Vision and
  Pattern Recognition}, 2016, pp. 4820--4828.

\bibitem{cordts2016cityscapes}
Marius Cordts, Mohamed Omran, Sebastian Ramos, Timo Rehfeld, Markus Enzweiler,
  Rodrigo Benenson, Uwe Franke, Stefan Roth, and Bernt Schiele,
\newblock ``The cityscapes dataset for semantic urban scene understanding,''
\newblock in {\em Proceedings of the IEEE Conference on Computer Vision and
  Pattern Recognition}, 2016, pp. 3213--3223.

\bibitem{xu2017end}
Huazhe Xu, Yang Gao, Fisher Yu, and Trevor Darrell,
\newblock ``End-to-end learning of driving models from large-scale video
  datasets,''
\newblock {\em arXiv preprint}, 2017.

\bibitem{wong2017segicp}
Jay~M Wong, Syler Wagner, Connor Lawson, Vincent Kee, Mitchell Hebert, Justin
  Rooney, Gian-Luca Mariottini, Rebecca Russell, Abraham Schneider, Rahul
  Chipalkatty, et~al.,
\newblock ``Segicp-dsr: Dense semantic scene reconstruction and registration,''
\newblock {\em arXiv preprint arXiv:1711.02216}, 2017.

\bibitem{paszke2016enet}
Adam Paszke, Abhishek Chaurasia, Sangpil Kim, and Eugenio Culurciello,
\newblock ``Enet: A deep neural network architecture for real-time semantic
  segmentation,''
\newblock {\em arXiv preprint arXiv:1606.02147}, 2016.

\bibitem{chaurasia2017linknet}
Abhishek Chaurasia and Eugenio Culurciello,
\newblock ``Linknet: Exploiting encoder representations for efficient semantic
  segmentation,''
\newblock {\em arXiv preprint arXiv:1707.03718}, 2017.

\bibitem{badrinarayanan2015segnet}
Vijay Badrinarayanan, Alex Kendall, and Roberto Cipolla,
\newblock ``Segnet: A deep convolutional encoder-decoder architecture for image
  segmentation,''
\newblock {\em arXiv preprint arXiv:1511.00561}, 2015.

\bibitem{long2015fully}
Jonathan Long, Evan Shelhamer, and Trevor Darrell,
\newblock ``Fully convolutional networks for semantic segmentation,''
\newblock in {\em Proceedings of the IEEE Conference on Computer Vision and
  Pattern Recognition}, 2015, pp. 3431--3440.

\bibitem{DBLP:journals/corr/RonnebergerFB15}
Olaf Ronneberger, Philipp Fischer, and Thomas Brox,
\newblock ``U-net: Convolutional networks for biomedical image segmentation,''
\newblock {\em CoRR}, vol. abs/1505.04597, 2015.

\bibitem{yu2015multi}
Fisher Yu and Vladlen Koltun,
\newblock ``Multi-scale context aggregation by dilated convolutions,''
\newblock {\em arXiv preprint arXiv:1511.07122}, 2015.

\bibitem{kingma2014adam}
Diederik Kingma and Jimmy Ba,
\newblock ``Adam: A method for stochastic optimization,''
\newblock {\em arXiv preprint arXiv:1412.6980}, 2014.

\bibitem{ioffe2015batch}
Sergey Ioffe and Christian Szegedy,
\newblock ``Batch normalization: Accelerating deep network training by reducing
  internal covariate shift,''
\newblock in {\em International Conference on Machine Learning}, 2015, pp.
  448--456.

\end{thebibliography}

\end{document}